\documentclass[aps,prx,reprint,preprintnumbers,superscriptaddress,nofootinbib,longbibliography,floatfix]{revtex4-2}
\pdfoutput=1
\usepackage{rotating}
\usepackage{array}
\usepackage{amsmath}
\usepackage[normalem]{ulem}
\usepackage{slashed}
\usepackage{booktabs}
\usepackage[pdftex,table]{xcolor}
\usepackage{units}
\usepackage{xfrac}
\usepackage{mathtools}
\usepackage{empheq}
\usepackage[]{units}
\usepackage{multirow}
\usepackage{amssymb}
\usepackage{url}
\usepackage{comment}
\usepackage{physics}
\usepackage{color,soul}
\usepackage{bbm}
\usepackage[caption=false]{subfig}

\usepackage{hyperref}
\hypersetup{
  colorlinks=true,
  citecolor=blue,
  linkcolor=blue,
  urlcolor=blue
}

\newcommand{\pt}{\mathrm{p_{T}}}

\begin{document}

\title{Online-compatible Unsupervised Non-resonant Anomaly Detection}

\author{Vinicius Mikuni}
\email{vmikuni@lbl.gov}
\affiliation{National Energy Research Scientific Computing Center, Berkeley Lab, Berkeley, CA 94720, USA}

\author{Benjamin Nachman}
\email{bpnachman@lbl.gov}
\affiliation{Physics Division, Lawrence Berkeley National Laboratory, Berkeley, CA 94720, USA}
\affiliation{Berkeley Institute for Data Science, University of California, Berkeley, CA 94720, USA}

\author{David Shih}
\email{shih@physics.rutgers.edu}
\affiliation{NHETC, Department of Physics \& Astronomy, Rutgers University, Piscataway, NJ 08854, USA}

\begin{abstract}
There is a growing need for anomaly detection methods that can broaden the search for new particles in a model-agnostic manner.  Most proposals for new methods focus exclusively on signal sensitivity.  However, it is not enough to select anomalous events - there must also be a strategy to provide context to the selected events.  We propose the first complete strategy for unsupervised detection of non-resonant anomalies that includes both signal sensitivity and a data-driven method for background estimation. Our technique is built out of two simultaneously-trained autoencoders that are forced to be decorrelated from each other.  This method can be deployed offline for non-resonant anomaly detection and is also the first complete online-compatible anomaly detection strategy.  We show that our method achieves excellent performance on a variety of signals prepared for the ADC2021 data challenge.

%In addition to providing a new strategy for offline anomaly detection, our new method is also online-compatible.  

%, but is not enough to design offline strategies.  At hadron colliders, nearly all events are discarded in real time - if the new physics is not saved, then offline methods will be unsuccessful.  Autoencoders are a class of unsupervised machine learning algorithms that can isolate anomalous regions of phase space with the low latency requirement of the trigger system at the Large Hadron Collider. However, an anomaly detection strategy requires achieving signal sensitivity and estimating the background.  So far, proposals for online-compatible anomaly detection have focused only on signal sensitivity.  We propose the first complete anomaly detection strategy that is online-compatible and can be used to both achieve signal sensitivity and is amenable to a data-driven background estimation.  Instead of training a single autoencoder, we propose training multiple autoencoders at the same time.  These networks are constructed to be independent of each other so that their reconstruction losses can be used with an ABCD background estimation strategy.  We show that our method achieves excellent performance on a variety of signals prepared for the ADC2021 data challenge.
\end{abstract}

\maketitle

%{\small
%\tableofcontents
%}

\section{Introduction}
\label{sec:intro}

Despite the compelling indirect evidence for new fundamental particles from astrophysical and other observations, no direct discoveries have been confirmed since the identification of the Higgs boson~\cite{ATLAS:2012yve,CMS:2012qbp}.  This means that the new physics is either rare, inaccessible, or we are looking in the wrong place for it.  This last possibility has motivated a new anomaly detection research program at particle colliders by which search strategies are constructed with less model dependence than previous approaches.  Many of these new methods employ modern machine learning to achieve broad sensitivity to unforeseen scenarios~\cite{Collins:2018epr,Collins:2019jip,Andreassen:2020nkr,Nachman:2020lpy,collaboration2020dijet,1815227,Stein:2020rou,Kasieczka:2021tew,Shih:2021kbt,Hallin:2021wme,Farina:2018fyg,Heimel:2018mkt,Roy:2019jae,Blance:2019ibf,Collins:2021nxn,Kahn:2021drv,Amram:2020ykb,Kasieczka:2021xcg,Hajer:2018kqm,DeSimone:2018efk,Mullin:2019mmh,1809.02977,Dillon:2019cqt,Romao:2019dvs,Romao:2020ojy,knapp2020adversarially,Cerri:2018anq,Govorkova:2021utb,Govorkova:2021hqu,1797846,1800445,Amram:2020ykb,Cheng:2020dal,pol2020anomaly,Atkinson:2021nlt,Khosa:2020qrz,Thaprasop:2020mzp,Alexander:2020mbx,aguilarsaavedra2020mass,vanBeekveld:2020txa,Park:2020pak,Faroughy:2020gas,Chakravarti:2021svb,Batson:2021agz,Blance:2021gcs,Bortolato:2021zic,Dillon:2021nxw,Finke:2021sdf,Aarrestad:2021oeb,Caron:2021wmq,Volkovich:2021txe,Ostdiek:2021bem,Fraser:2021lxm,DAgnolo:2018cun,DAgnolo:2019vbw,Dorigo:2021iyy,Aguilar-Saavedra:2017rzt,Mikuni:2020qds} (list from Ref.~\cite{Feickert:2021ajf}). 

A complete anomaly detection algorithm is required to have two attributes: it should be sensitive to anomalous events and it should be possible to estimate the rate of Standard Model (SM) events that are labeled as anomalous (false positive rate)~\cite{Nachman:2020lpy}.  Complete anomaly detection methods have so far primarily focused on {\it resonant} anomalies, where data sidebands can be used as reference samples to both construct signal-sensitive classifiers and to estimate the SM background~\cite{Collins:2018epr,Collins:2019jip,Andreassen:2020nkr,Nachman:2020lpy,collaboration2020dijet,Amram:2020ykb,1815227,Stein:2020rou,Kahn:2021drv,Kasieczka:2021tew,Hallin:2021wme,Shih:2021kbt}. %{\bf A thought: put in a sentence or two about how CATHODE achieves optimal performance and so the resonant anomaly detection is a mostly solved problem now. The new frontier is nonresonant anomaly detection.}

Much less well-explored so far has been complete anomaly detection methods for {\it non-resonant} anomalies. One widely studied approach based on unsupervised learning that does not require the new physics to be resonant is the autoencoder~\cite{Farina:2018fyg,Heimel:2018mkt,Roy:2019jae,Cerri:2018anq,Blance:2019ibf,Hajer:2018kqm,DeSimone:2018efk,Mullin:2019mmh,1809.02977,Dillon:2019cqt,Romao:2019dvs,Romao:2020ojy,knapp2020adversarially,1797846,1800445,Amram:2020ykb,Cheng:2020dal,Khosa:2020qrz,Thaprasop:2020mzp,Alexander:2020mbx,aguilarsaavedra2020mass,pol2020anomaly,vanBeekveld:2020txa,Park:2020pak,Faroughy:2020gas,Kasieczka:2021xcg,Chakravarti:2021svb,Batson:2021agz,Blance:2021gcs,Bortolato:2021zic,Collins:2021nxn,Dillon:2021nxw,Finke:2021sdf,Atkinson:2021nlt,Kahn:2021drv,Aarrestad:2021oeb,Caron:2021wmq,Govorkova:2021hqu,Volkovich:2021txe,Govorkova:2021utb,Ostdiek:2021bem,Fraser:2021lxm}.
%Cite some ML papers?  Maybe this one is the first? https://dl.acm.org/doi/pdf/10.1145/2689746.2689747
The idea is to build models for compressing and uncompressing events, trained directly on the (mostly background) data.  Events that have a low probability density tend to be poorly reconstructed when compressing and uncompressing compared with events that have a relatively higher probability density.  If anomalous events are located in regions of low data probability density, then the reconstruction quality can be used as an anomaly score. 

 However, autoencoders by themselves are not a complete anomaly detection algorithm - they provide a method for achieving signal sensitivity, but they do not have a natural background estimation component.  
 %For offline resonant anomaly detection, one could combine reconstruction quality with sidebands to form a complete strategy.   
 In the non-resonant case, one could compare the spectrum of anomalous events with background-only simulations, but this requires an excellent model of the background.  Given that we expect the unexpected to occur in regions that are poorly modeled, this is unlikely to be a viable strategy in general.

 In this paper, we introduce a new method for detecting non-resonant anomalies, based on autoencoders, that is complete in the sense that it includes both signal sensitivity and simulation-free background estimation. Instead of constructing one autoencoder, we advocate for training two or more autoencoders.  The set of autoencoders are trained to be as independent of each other as possible.
While many methods for decorrelating neural networks exist~\cite{Louppe:2016ylz,Dolen:2016kst,Moult:2017okx,Stevens:2013dya,Shimmin:2017mfk,Bradshaw:2019ipy,ATL-PHYS-PUB-2018-014,DiscoFever,Xia:2018kgd,Englert:2018cfo,Wunsch:2019qbo,Rogozhnikov:2014zea,10.1088/2632-2153/ab9023,clavijo2020adversarial,Kasieczka:2020pil,Kitouni:2020xgb,Ghosh:2021hrh} and could be used here, we chose to employ the DisCo decorrelation method first developed in Ref.~\cite{DiscoFever} and explored for simultaneous background estimation in Ref.~\cite{Kasieczka:2020pil}.
%We accomplish this using the DisCo decorrelation method first developed in Ref.~\cite{DiscoFever} and explored for simultaneous background estimation in Ref.~\cite{Kasieczka:2020pil}.  %BPN: this is a repeat for what comes later.  I don't think we need to mention this level of detail in the introduction.
Events are labeled as anomalous if the reconstruction quality is poor for all autoencoders.  Events labeled as anomalous by one, but not all, of the autoencoders provide the context needed to estimate the Standard Model background in a model independent way, via the ABCD method.% {\bf Describe ADC2021 dataset and highlights?}

An additional benefit of our method is that it can be run equally well online or offline; indeed this forms a second major motivation for our work. Typically, a key assumption is that anomalous events will be saved by the detectors for offline analysis. Due to the immense data rate at the Large Hadron Collider (LHC), it is not possible to save every collision event for offline processing.  Instead, a system of triggers are used to save interesting events~\cite{ATLAS:2020esi,CMS:2016ngn}.  The definition of interesting is model dependent and therefore the new physics may be thrown away in real time.  It is therefore of utmost importance to design model independent strategies for saving anomalous events. 

%Autoencoders can be run online because they do not require comparing data to a reference sample

Autoencoders can be run online because they do not require comparing data to a reference sample~\cite{knapp2020adversarially,Cerri:2018anq,Govorkova:2021utb,Govorkova:2021hqu}. However, no autoencoder-based trigger proposal so far has been complete in the sense introduced above.  Many conventional triggers are complemented by \textit{support} triggers which provide the context needed for data-driven background estimation offline.  Our method provides the first complete anomaly detection strategy in a similar way to these conventional methods.  By using two decorrelated autoencoders, we can trigger on potentially anomalous events and then additionally save (at a reduced rate) anti-tagged events in a way that background estimation is possible offline.

This paper is organized as follows.  First, we introduce the technique of decorrelated autoencoders in Sec.~\ref{sec:method}.  Numerical results with the ADC2021 dataset are presented in Sec.~\ref{sec:results}.  By definition, this demonstration highlights an offline application of our approach.  Section~\ref{sec:online} provides a discussion about the online-compatibility of our technique for experimental integration online.  The paper ends with conclusions and outlook in Sec.~\ref{sec:conclusions}.

\section{Decorrelated Autoencoders}
\label{sec:method}

A vanilla autoencoder is a composition of two functions, an encoder $g$ and a decoder $f$.  These two functions are parameterized as neural networks and are optimized to minimize the reconstruction loss:

\begin{align}
\label{eq:auto}
L[f,g]=\sum_i (f(g(x_i))-x_i)^2\,,
\end{align}
where $x\in\mathbb{R}^n$, $g:\mathbb{R}^n\rightarrow\mathbb{R}^m$, and $f:\mathbb{R}^m\rightarrow\mathbb{R}^n$.  In order to encourage compression, the latent space dimension is chosen such that $m<n$.  A popular variation on this setup is the variational autoencoder~\cite{kingma2014autoencoding,Kingma2019}, whereby the encoding and decoding are probabilistic and the latent space has well-defined statistical properties.  The methods proposed here are compatible with variational autoencoders, and while preliminary studies indicate that the results are similar, we leave a full exploration to future work.
%but we leave this combination for future work.

Instead of training a single autoencoder as in Eq.~\ref{eq:auto}, we propose to train two (or more) {\it statistically independent} autoencoders at the same time, in order to enable data-driven background estimation. Following \cite{DiscoFever,Kasieczka:2020pil}, we achieve the decorrelation of the autoencoders by including in the training a regularizer term based on the distance correlation (DisCo) measure of statistical dependence. Focusing on the case of two autoencoders $(f_1,g_1)$ and $(f_2,g_2)$ for simplicity, we consider the following loss function: 
\begin{align}
\label{eq:autodec}\nonumber
L[f_1,f_2,g_1,g_2]=&\sum_i R_1(x_i)^2+\sum_i R_2(x_i)^2\\
&+\lambda\, \text{DisCo}^2[R_1(X),R_2(X)]\,,
\end{align}
where $R_i(x)=(f_i(g_i(x))-x)^2$, $\lambda > 0$ is a hyperparameter, and DisCo is the distance correlation~\cite{szekely2007, szekely2009, SzeKely:2013:DCT:2486206.2486394,szekely2014}.  DisCo is between $0$ and $1$ and is zero if and only if its arguments are independent.  The capital $X$ is used in the last term of Eq.~\ref{eq:autodec} to indicate that the distance correlation is computed at the level of a batch of examples $x$, which are realizations of the random variable $X$.    Given autoencoders trained via Eq.~\ref{eq:autodec}, we can define counts $N_{\lessgtr,\lessgtr}(\vec{c})=\sum_i\mathbb{I}[R_1(x_i)\lessgtr c_1]\,\mathbb{I}[R_2(x_i)\lessgtr c_2]$, where $\vec{c}=(c_1,c_2)$ are given thresholds and $\mathbb{I}[\cdot]$ is the indicator function that is zero when its argument is false and one otherwise.  The signal sensitive region is $N_{>,>}(\vec{c})$ and the other three regions can be used to estimate the background: 

\begin{align}
\label{eq:abcd}
N_{>,>}^\text{predicted}(\vec{c})&=\frac{N_{>,<}(\vec{c})N_{<,>}(\vec{c})}{N_{<,<}(\vec{c})}\,.
\end{align}
Equation~\ref{eq:abcd} is known as the ABCD method and the $N_{>,>}(\vec{c})$ is exactly the background in the signal-sensitive region if there are enough events and if the two dimensions are effective at rejecting the background.

%\section{Anomaly detection at trigger level}
\section{Empirical Results}
\label{sec:results}

The performance of the double autoencoder and decorrelation strategy is tested on the ADC2021 dataset, which was created for unsupervised anomaly detection~\cite{40mhz,Govorkova:2021hqu}.  In the dataset, proton-proton collisions at the LHC are simulated at center-of-mass energy of 13~TeV. Collision events are required to contain at least one electron ($e$) or muon ($\mu$) with transverse momenta $\pt > 23$~GeV.  A set of various Standard Model processes are generated with \textsc{Pythia} 8.240 generator~\cite{Sjostrand:2006za,Sjostrand:2014zea} with detector response modeled by \textsc{Delphes} 3.3.2~\cite{deFavereau:2013fsa,Selvaggi:2014mya,Mertens:2015kba} using the Phase-II CMS detector card.  %Background SM processes considered include inclusive $W$ boson production ($59.2\%$ of the dataset), inclusive $Z$ boson production ($6.7\%$ of the dataset), top quark pair production ($0.3\%$ of the dataset), and Quantum Chromodynamics (QCD) multijet production ($33.8\%$ of the dataset). 
During the training, 2 million events are used while results are reported using an independent validation set containing 800k SM events.

Four benchmark scenarios containing new physics processes are used to evaluate the performance of the algorithm: a leptoquark (LQ) with 80~GeV mass decaying to a $b$-quark and a $\tau$ lepton, a neutral scalar boson ($A$) of 50~GeV mass decaying to a pair of off-shell $Z$ bosons, which in turn are forced to decay to leptons ($A\rightarrow 4l$), a scalar boson $h^0$ of 60~GeV mass decaying to a pair of $\tau$ leptons ($h^0\rightarrow \tau\tau$), and a charged scalar boson $h^\pm$ with 60~GeV mass, decaying to a $\tau$ lepton and a neutrino ($h^\pm\rightarrow\tau\nu$). In the performance evaluation, each new physics scenario is considered independently, with total amount of events fixed to 0.1\% of the total sample size.

The autoencoders are trained on a sample of pure background events.  In practice, this corresponds to the case of training on simulation and testing on data.  Differences between data and simulation (which are not modeled or taken into account in the ADC2021 dataset used here) may degrade the autoencoder performance if background data events are not reconstructed as well by the autoencoder as background simulation events.  Fortunately, it is well-known from previous studies (see e.g.~\cite{Farina:2018fyg,Heimel:2018mkt,Cerri:2018anq}) that autoencoder training is highly insensitive to low amounts of signal contamination.  This means that autoencoders can be trained directly on data with a small amount of signal contamination without a significant change in the learned neural networks.  We have explicitly verified this in the case of the decorrelated autoencoders using the $A\rightarrow 4l$ signal.

Each autoencoder architecture is built using deep neural networks containing five fully connected layers. The encoders have 256, 128, 64, 32, and 5 hidden nodes, while the decoder is simply the mirrored version of the encoder. The inputs given to the training are the four-momenta of jets~\cite{Cacciari:2011ma,Cacciari:2005hq} and leptons in ($\pt$, $\eta$, $\phi$, m) coordinates. %, which are then regressed in the output using the mean square error loss. 
Only the first (sorted by $\pt$) four muons, four electrons, and 10 jets in the event are kept with zero padding if fewer objects are present.
% If fewer objects are present the event is zero-padded while events with more particles are truncated after ordering particles by decreasing $\pt$ values.
The implementation is carried out with \textsc{Tensorflow}~\cite{tensorflow} optimized with the \textsc{Adam}~\cite{adam}. Even though all new physics scenarios considered here contain a mass resonance, no invariant mass information is directly used in the training process. The $\lambda$ parameter from Eq.~\ref{eq:autodec} is fixed to 100 and training batch size fixed to 10k to improve the decorrelation performance. The double autoencoder structure is then trained for a total of 1000 epochs, or stopped if the overall training loss does not improve in an independent testing set for 10 consecutive epochs. The complete model uses 230k trainable weights with a total of 460k floating point operations.  The neural network architecture and training procedure were not extensively optimized, due in part to the unsupervised nature of this task.

\begin{figure}[h!]
\centering
\includegraphics[width=0.4\textwidth]{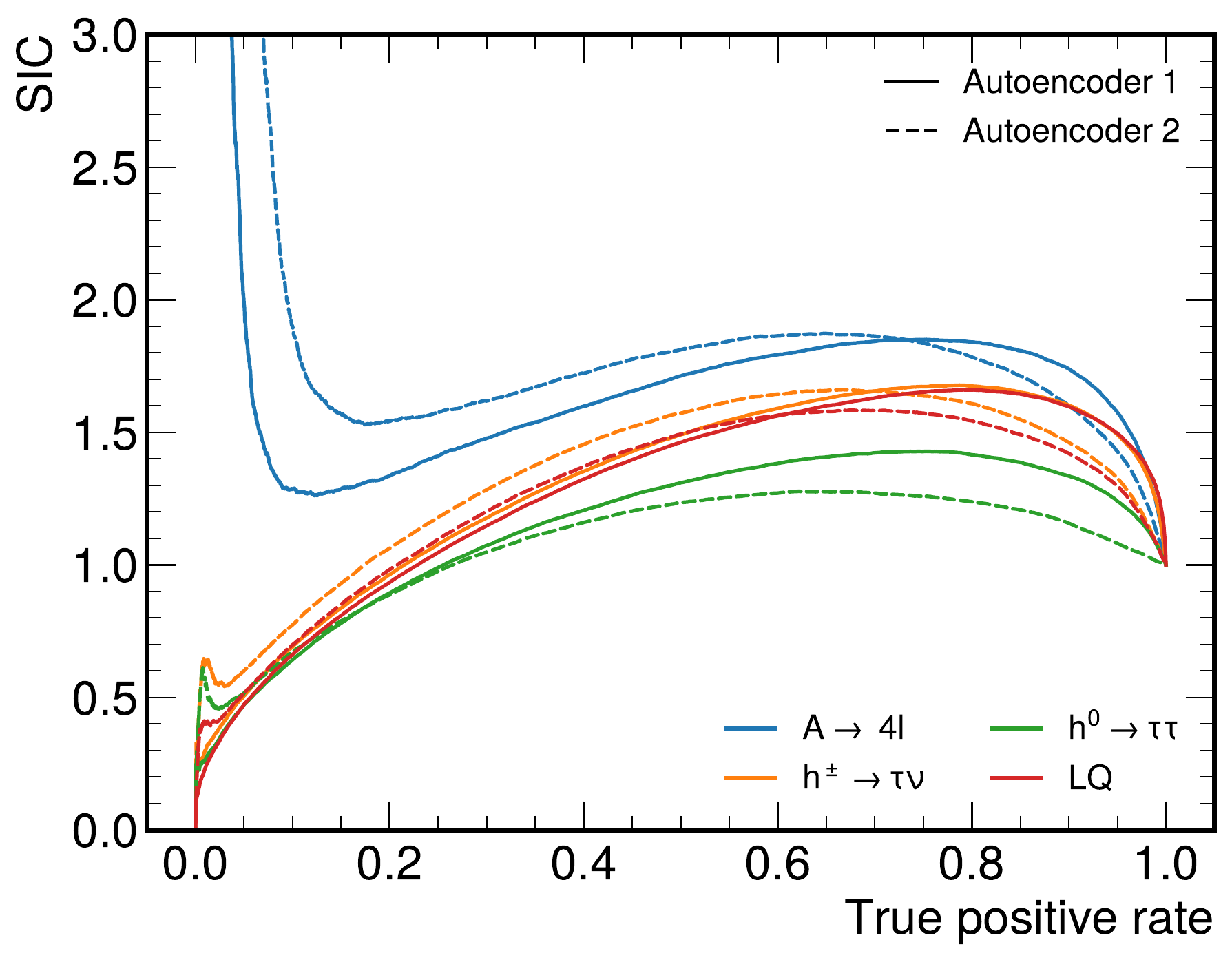}
\includegraphics[width=0.4\textwidth]{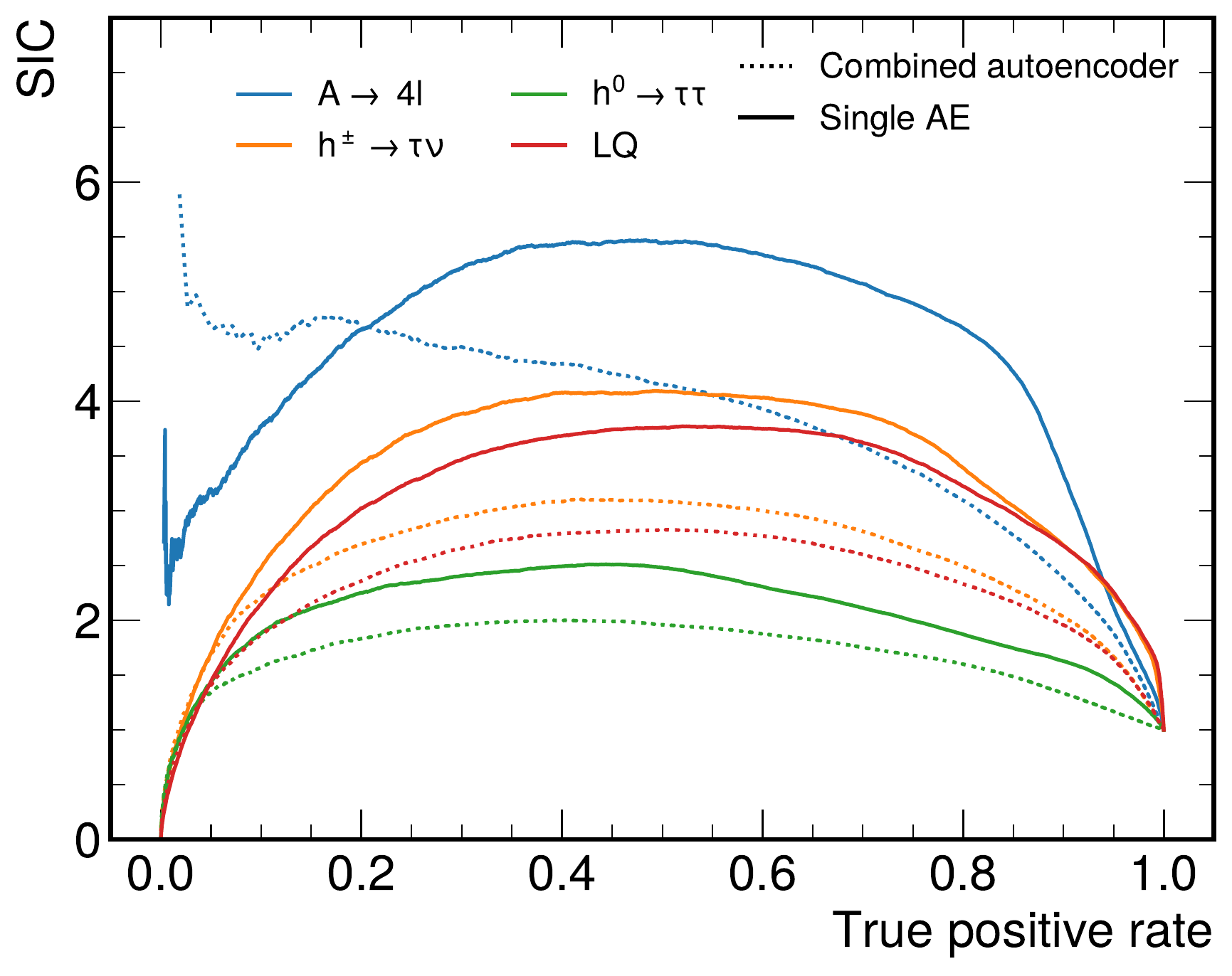}
%\includegraphics[width=0.4\textwidth]{figures/roc_comparison_double.pdf}

% \caption{Receiver operating characteristic (top) and  significance improvement characteristic (bottom) curves for each new physics benchmark scenario and for each separate autoencoder (dashed and dashed-dotted lines). A single autoencoder trained without the ecorrelation loss is shown as a solid line. The combined response for both autoencoders is calculated by selecting the output threshold that yields the same fake rate for both autoencoders, shown as a dotted line.}
\caption{Top: significance improvement characteristic (SIC) curves of the individual, decorrelated autoencoders, for each new physics benchmark scenario. Bottom: SIC curves for the ``diagonal" combination of both decorrelated autoencoders, compared with that of a single autoencoder (obviously trained without any decorrelation). In both panels, the SIC curves are cut off at lower true positive rates due to low background yields.
% and for each separate autoencoder (top).  A hsingle autoencoder trained without the decorrelation loss is compared to the performance of the combined autoencoder (bottom). The combined response for both autoencoders is calculated by selecting the output threshold that yields the same fake rate.
}
\label{fig:roc_sic}
\end{figure}

The performance of each autoencoder for anomaly detection is assessed by using the reconstruction loss as the main discriminator. The
significance improvement characteristic (SIC) curves are built for each new physics scenario shown in Fig.~\ref{fig:roc_sic}. The comparison with a single autoencoder trained without the decorrelation loss is also shown. We also show in Fig.~\ref{fig:roc_sic} the combined performance of both autoencoders, where a ``diagonal" cut that yields the same SM background efficiency for each autoencoder is employed. We see that the signal sensitivity of the combined autoencoders is greater than (in fact roughly double) that of each autoencoder individually, indicating that the decorrelation was successful and each autoencoder learned something independent about the BSM anomalies in question. The SIC distribution for different combinations of thresholds in each autoencoder is shown in App.~\ref{app:sic_curves}.

\begin{figure}[h!]
\centering
\includegraphics[width=0.45\textwidth]{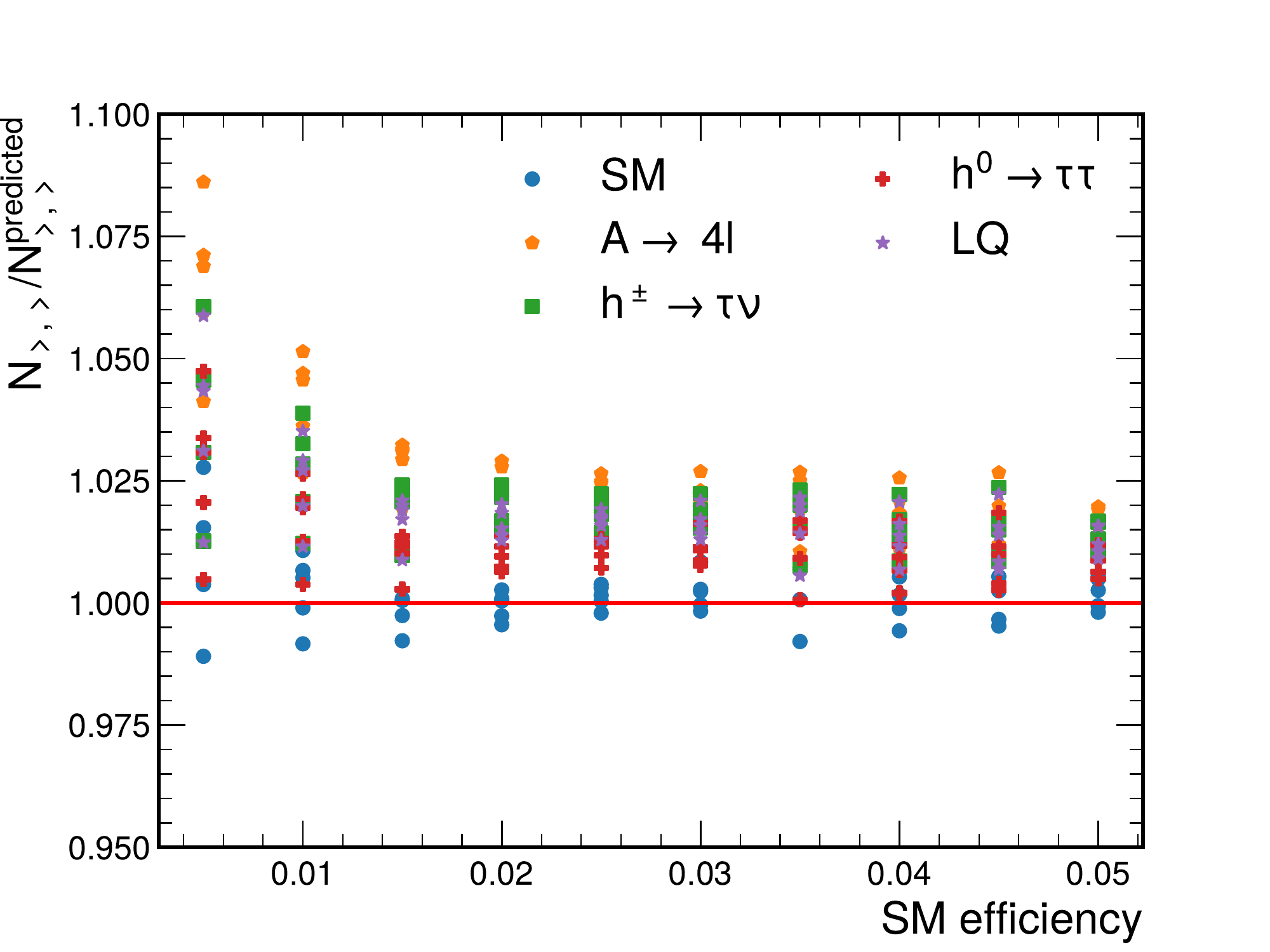}
\caption{Closure test of the ABCD estimation method for different SM efficiencies and benchmark scenarios. Different selection combinations yielding the same background efficiency are shown as independent entries.}
\label{fig:closure}
\end{figure}

The independence between reconstruction losses is more fully validated by estimating the difference between the background predicted using the ABCD method (Eq.~\ref{eq:abcd}) and the real number of background events in the region of interest. The ratio between the two quantities is shown in Fig.~\ref{fig:closure}. Multiple choices of $\vec{c}$ yielding the same SM efficiency are tested (represented as multiple entries in Fig.~\ref{fig:closure}) for samples containing only SM processes (purple) and mixtures of SM and a new physics process. At lower SM efficiencies, departures from unity (overestimated background~\cite{Kasieczka:2020pil}) are observed in all mixed samples while the highest variation for a sample containing only SM events is 2.5\%, compatible the statistical uncertainty of the sample.

%{\bf So background tends to be overestimated by our method? I forget, is there a reason for that? In any case, if it's a systematic effect and not just random, then this is good for avoiding false discoveries}
%Yes, expect it to be overestimated - see Eq. 2.10 in https://arxiv.org/pdf/2007.14400.pdf

\begin{figure}[h!]
\centering
\includegraphics[width=0.45\textwidth]{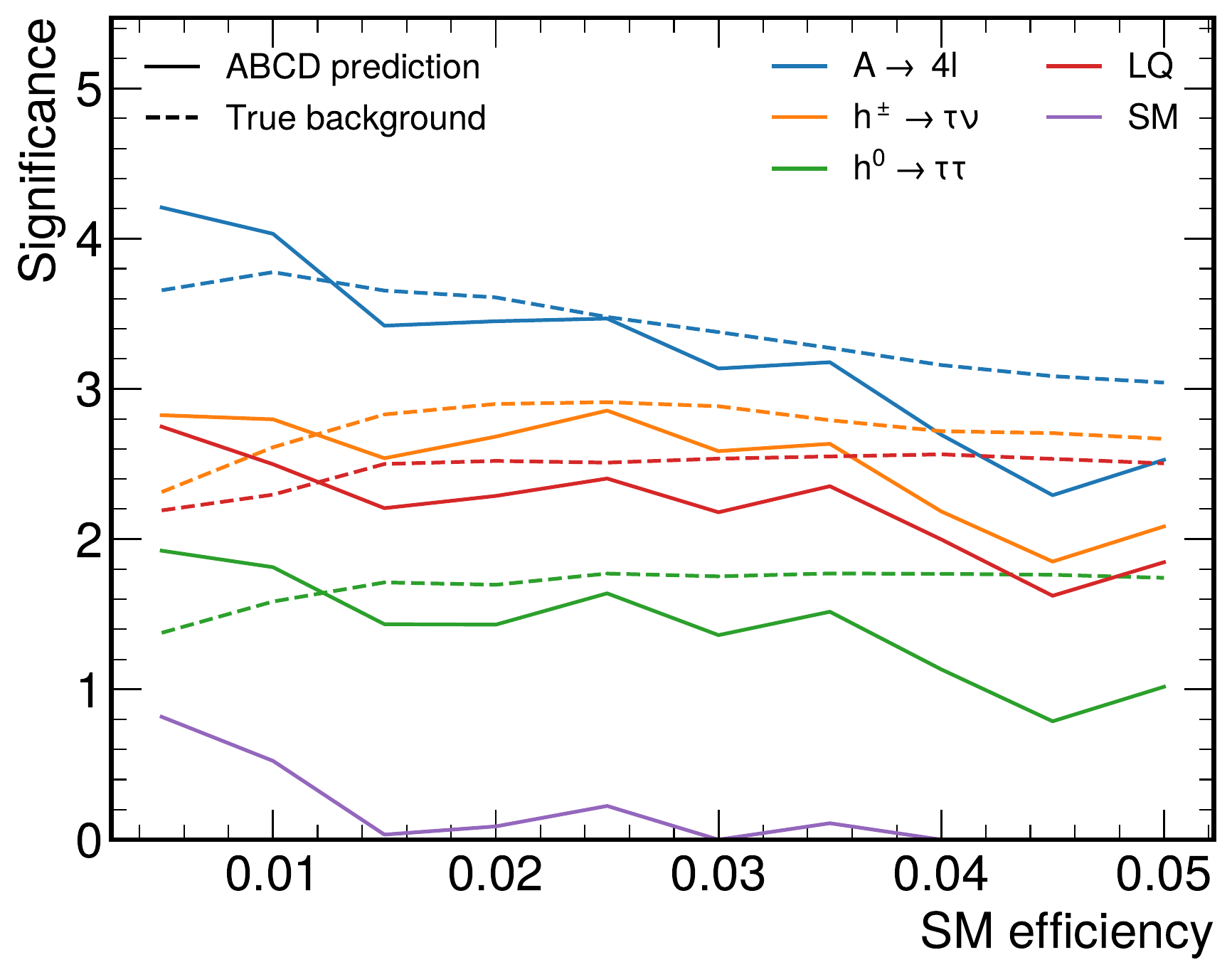}
\includegraphics[width=0.45\textwidth]{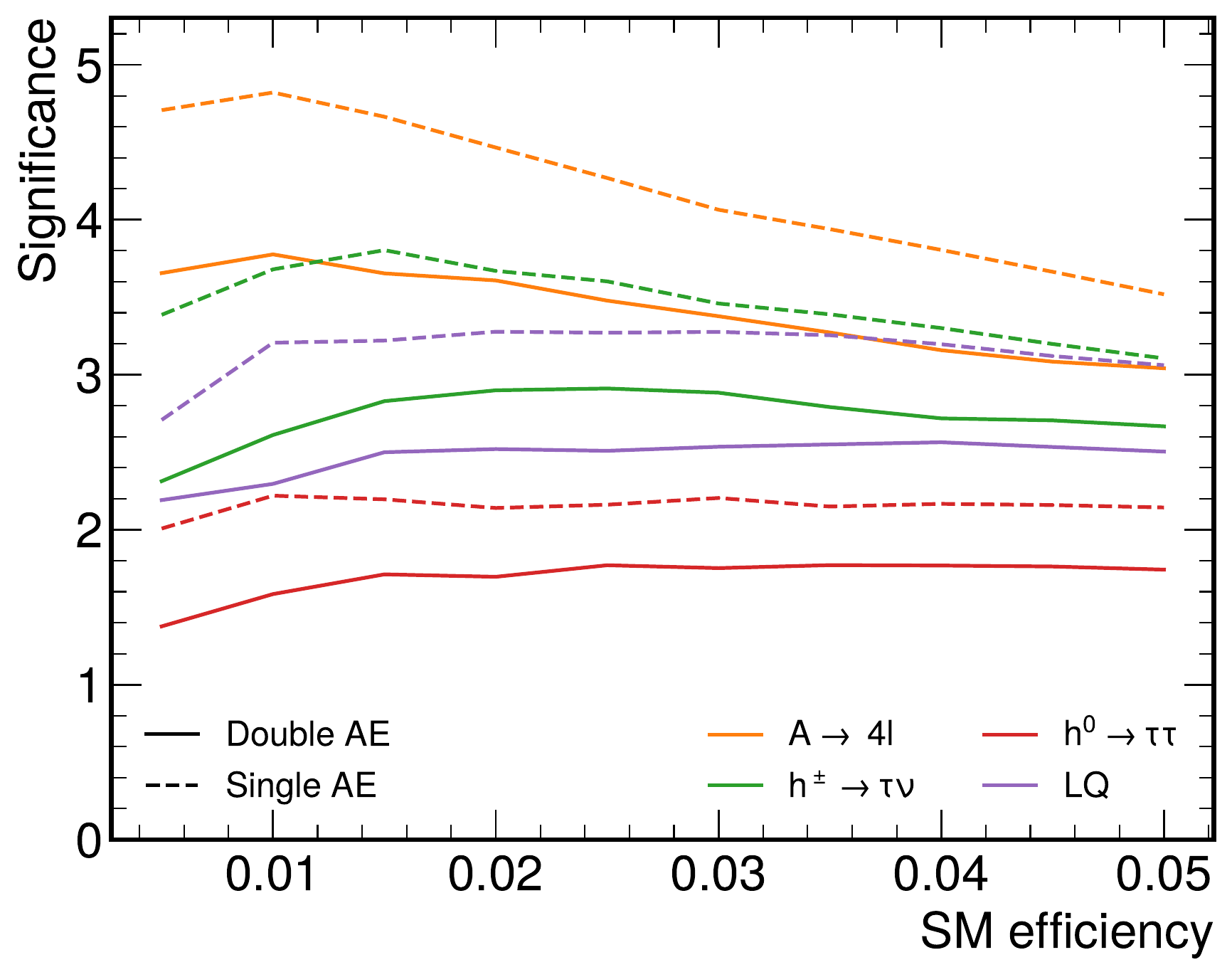}
\caption{Signal significance for each benchmark scenario (top) when the ABCD method is used to predict the background level (solid lines) compared to the real significance value (dashed lines). In the bottom panel, the comparison of the significance between a single autoencoder (dashed lines) and the double autoencoder (solid lines) is shown. In this case, no background estimation method is used.}
\label{fig:sig}
\end{figure}

These differences can also be quantified in terms of the signal significance for each benchmark process by comparing the observed and predicted number of background events from the ABCD method. Given $N$ observations in the region of interest with predicted number of background events $B$, the significance is defined as $\frac{N-B}{\sqrt{N}}$ if $N - B >0$ and 0 otherwise. With the initial signal fraction fixed, the total sample significance is around 0.8 prior to the application of the method. As pointed out in Ref.~\cite{Kasieczka:2020pil}, unaccounted contamination from the signal of interest in the ABCD sidebands may result in different significance values when compared to the correct estimation of the background. While this issue can be accounted when performing model specific exclusion limits, we also show in Fig.~\ref{fig:sig} (top) the significance obtained using the ABCD method with and without correcting the number of background events. To avoid fine tuning, the threshold applied to each autoencoder reconstruction loss is the one where both autoencoders have the same SM rejection efficiency.

In all new physics benchmark scenarios, the uncorrected significance for SM efficiencies above 1.5\% is lower than the corrected for SM efficiencies. Nevertheless, all new physics scenarios show significance between 1 to 4 while the SM only sample has a maximum deviation below 1. We have also probed the stability of the method by performing five independent trainings with different random weight initialization. The standard variation of the average significance was below 6\% for all benchmark scenarios tested.

The additional distance correlation loss leads to increased reconstruction loss in the background training sample, resulting in decreased performance compared to a single autoencoder training. This difference is illustrated in Fig.~\ref{fig:sig} (bottom) where the significance is compared with the values obtained from training a single autoencoder with same network architecture. Since the ABCD method is only applicable in the double autoencoder case, no background estimation method is used in the comparison. In all cases, the difference in significance of the double and single autoencoders is less than 30\%. While the single autoencoder consistently outperforms the double autoencoder, the lack of dedicated background estimation might lead to unattainable performances when applied to real particle collisions.

% \section{Gaussian Example}
% \label{sec:gaussian}

% The data are generated as $X_2 \sim \mathcal{N}(\mu,1.5), X_1\sim\mathcal{N}(\mu,1.5), X_0=\rho X_1+(1-\rho^2) Z$ where $Z\sim\mathcal{N}(\mu,1.5)$.   The random variables $X_2,X_1,Z$ are independent.   For the background, $\rho=0.8$ and $\mu=0$; for the signal, $\rho=0$ and $\mu=2$.  We generate 2 million events.

% The results are shown in Fig.~\ref{fig:toy}.  The functions $f_i$ and $g_i$ each have one hidden layer with 50 nodes each.  The latent space is two-dimensional.  The hyperparameter $\lambda=10$.

% \begin{figure}[h!]
% \centering
% \includegraphics[width=0.45\textwidth]{figures/Toy_decAE.pdf}
% \caption{To be filled in.}
% \label{fig:toy}
% \end{figure}

\section{Anomaly Detection Online}
\label{sec:online}

%Our decorrelated autoencoder protocol as described above is a first completely data-driven approach for non-resonant anomaly detection offline. 

The discussion so far has demonstrated that the decorrelated autoencoder protocol is an effective tool for simulation-free, non-resonant anomaly detection.  This section briefly describes how this technique is also online compatible. 
%Our decorrelated autoencoder is online compatible because it provides a way to simultaneously achieve signal sensitivity and estimate the background.  
We envision that in an actual trigger system, we would save all events in the signal sensitive region defined by the two autoencoders and then save a random fraction (`prescale') of events in the three other regions for offline background estimation (similar to existing `support triggers' for certain background processes).  The prescale would be set so that the statistical uncertainty on the background prediction is smaller than the statistical uncertainty from events in the signal region.  If the SM efficiency in the signal region is $\epsilon$, the trigger rate would scale approximately as $4\epsilon$, including events saved from the background-dominated regions.  The autoencoders themselves could be trained directly on data.  These data could be from a previous run or from earlier in a given run.   We note that this is the first complete online compatible anomaly detection protocol to be proposed - previous proposals have used single autoencoders and do not come with a method for estimating the background.  %~\cite{knapp2020adversarially,Cerri:2018anq,Govorkova:2021utb,Govorkova:2021hqu}.%maybe don't cite them here in order to directly criticize them?  We have the citation block earlier in the paper

Moreover, each of our autoencoders is built using only a set of fully connected layers to allow for a memory and time efficient implementation.  There have been many recent demonstrations of ultra low-latency implementations of these and related architectures on Field Programable Gate Arrays (FPGAs)~\cite{Duarte:2018ite,Coelho:2020zfu,Aarrestad:2021zos,Heintz:2020soy,Govorkova:2021utb,John:2020sak,Hong:2021snb}. For studies with more computational resources available, the baseline performance of each autoencoder may be enhanced using more complex reconstruction strategies, as studied in \cite{Cheng:2020dal,pol2020anomaly,Atkinson:2021nlt,Collins:2021pld,Orzari:2021suh}.
%\cite{Cerri:2018anq,Cheng:2020dal,pol2020anomaly,Atkinson:2021nlt,Collins:2021pld,Orzari:2021suh}

The ADC2021 community challenge dataset was used in part because it was created for the purpose of developing online methods~\cite{40mhz,Govorkova:2021hqu} as summarized by the challenge title: \textit{Unsupervised New Physics detection at 40 MHz}.   However, there are some features of this dataset which limit direct connection to online algorithms.  For example, ATLAS and CMS have single lepton triggers that would likely save all of the challenge events for offline processing.  Figure~\ref{fig:sig} indicates that our decorrelated autoencoder trigger reduces the bandwidth by nearly two orders of magnitude.  This is not necessarily relevant for the lepton-triggered data, but it is a common reduction for dedicated triggers. Another issue is, as we have already noted above, that the ADC2021 dataset does not distinguish between ``data" and ``simulation"; this could be an issue for machine learning methods, which generally require a representative sample for the training data. Expanding the online challenge to other datasets would be interesting for the future.   %It is highly non-trivial to propose well-motivated signals that would not be collected with existing triggers and so the ADC2021 dataset provides an important benchmark for a variety of studies.

%We use this dataset because it is a community data challenge created for exactly the purpose of online anomaly detection research and development.  However, we note that ATLAS and CMS have single lepton triggers that would likely save all of these events for offline processing. Expanding the online challenge to other datasets would be interesting for the future.

%Therefore we view this method as trainable directly on data {\bf although then wouldn't we have had to trigger on this data previously? then how is this any different than needing to save sidebands for the resonant anomaly detectors?}

\section{Conclusions and Outlook}
\label{sec:conclusions}

We have proposed a first complete online-compatible unsupervised non-resonant anomaly detection method that achieves signal sensitivity and can be used to estimate the SM background\footnote{The scripts used to produce the results shown in this work are available at: \url{https://github.com/ViniciusMikuni/DoubleAE}}.  Autoencoders, a popular choice of anomaly detection algorithm, are used to identify anomalous events through a reconstruction loss. We advocate for the combination of two or more autoencoders that are trained simultaneously while a distance correlation  (DisCo) regularizer term is added to make their reconstruction losses statistically independent. In this strategy, the background from SM events can be estimated with the ABCD method. In the absence of new physics, the method shows a good agreement between the predicted and observed amount of background events. In the presence of new physics, the signal significance varies between 1 and 4 for multiple new physics scenarios with initial contribution amounting to only 0.1\% of all events.  
%While the significance of multiple autoencoders is reduced compared to a single autoencoder, the reconstructed loss threshold can be optimized for each benchmark of interest. 
Given that our method is architecture agnostic, it can be readily generalized for other anomaly detection methods whose output is an anomalous score capable of discerning new physics scenarios from background events.

\section*{Acknowledgments}

We thank Barry Dillon and Gregor Kasieczka for feedback on the manuscript.  VM and BN are supported by the U.S. Department of Energy (DOE), Office of Science under contract DE-AC02-05CH11231. The work of DS was supported by DOE grant DOE-SC0010008. 

\bibliography{HEPML,other}

\clearpage
\appendix

\section{Significance improvement characteristic for different selection thresholds}
\label{app:sic_curves}

\begin{figure}[t]
\centering
\includegraphics[width=0.20\textwidth]{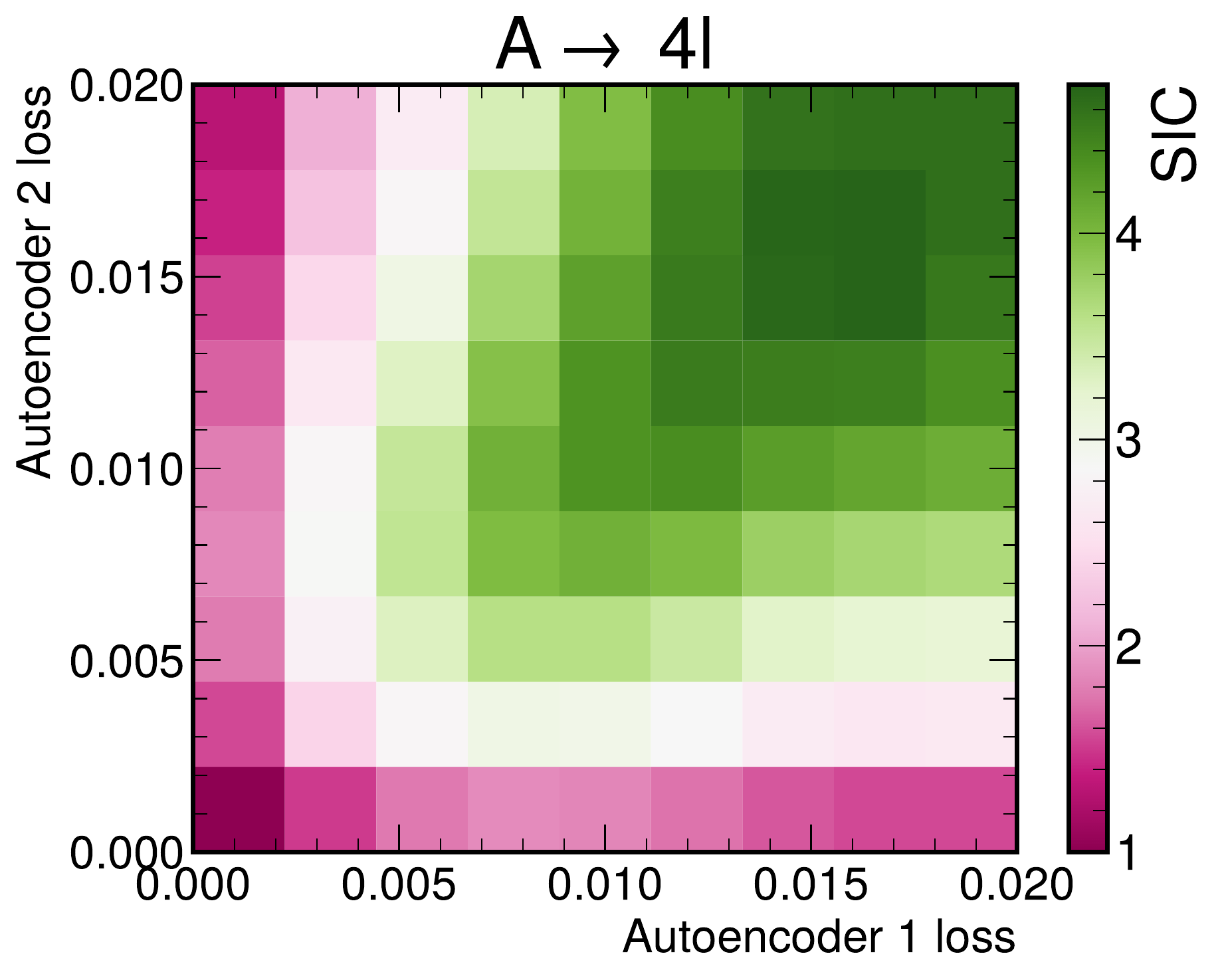}
\includegraphics[width=0.20\textwidth]{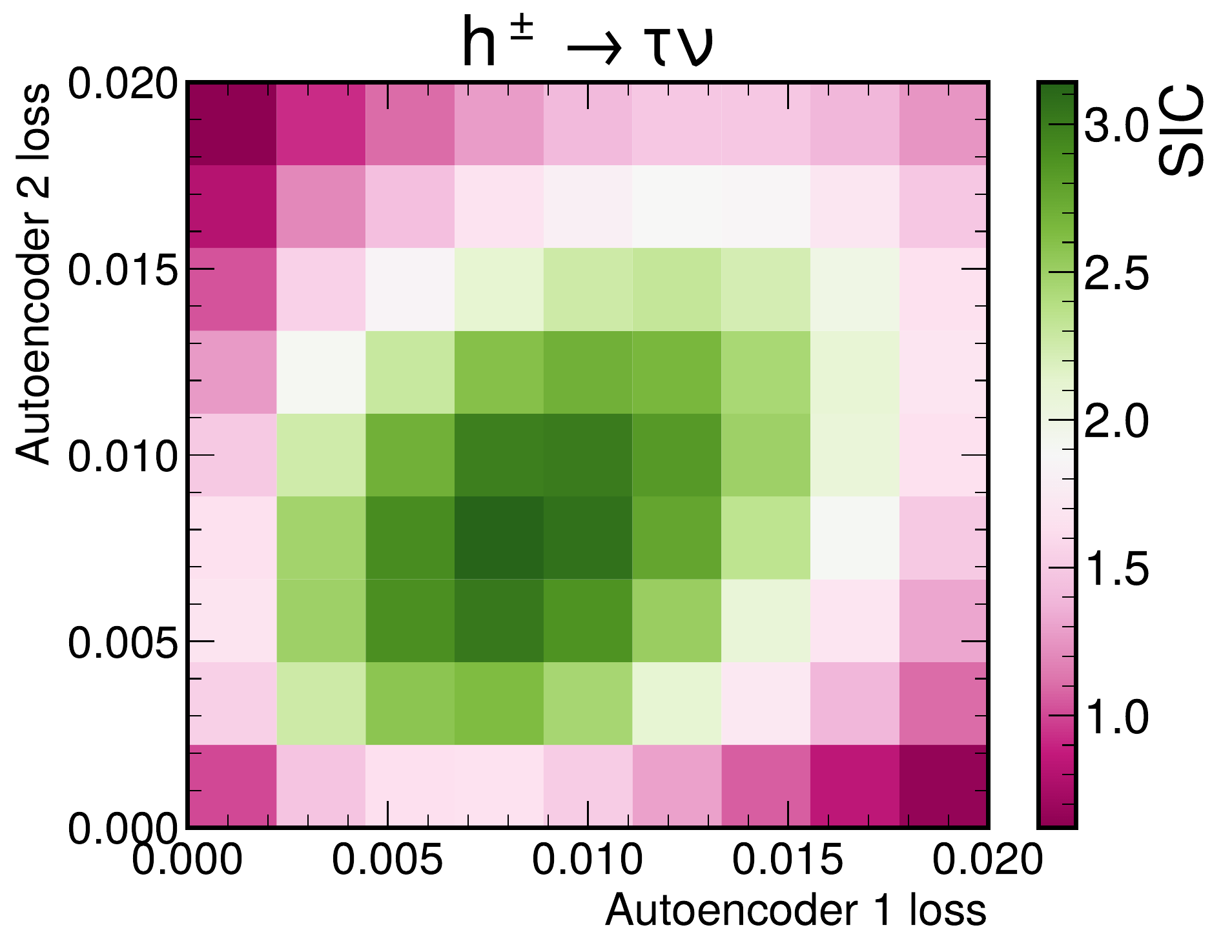}
\includegraphics[width=0.20\textwidth]{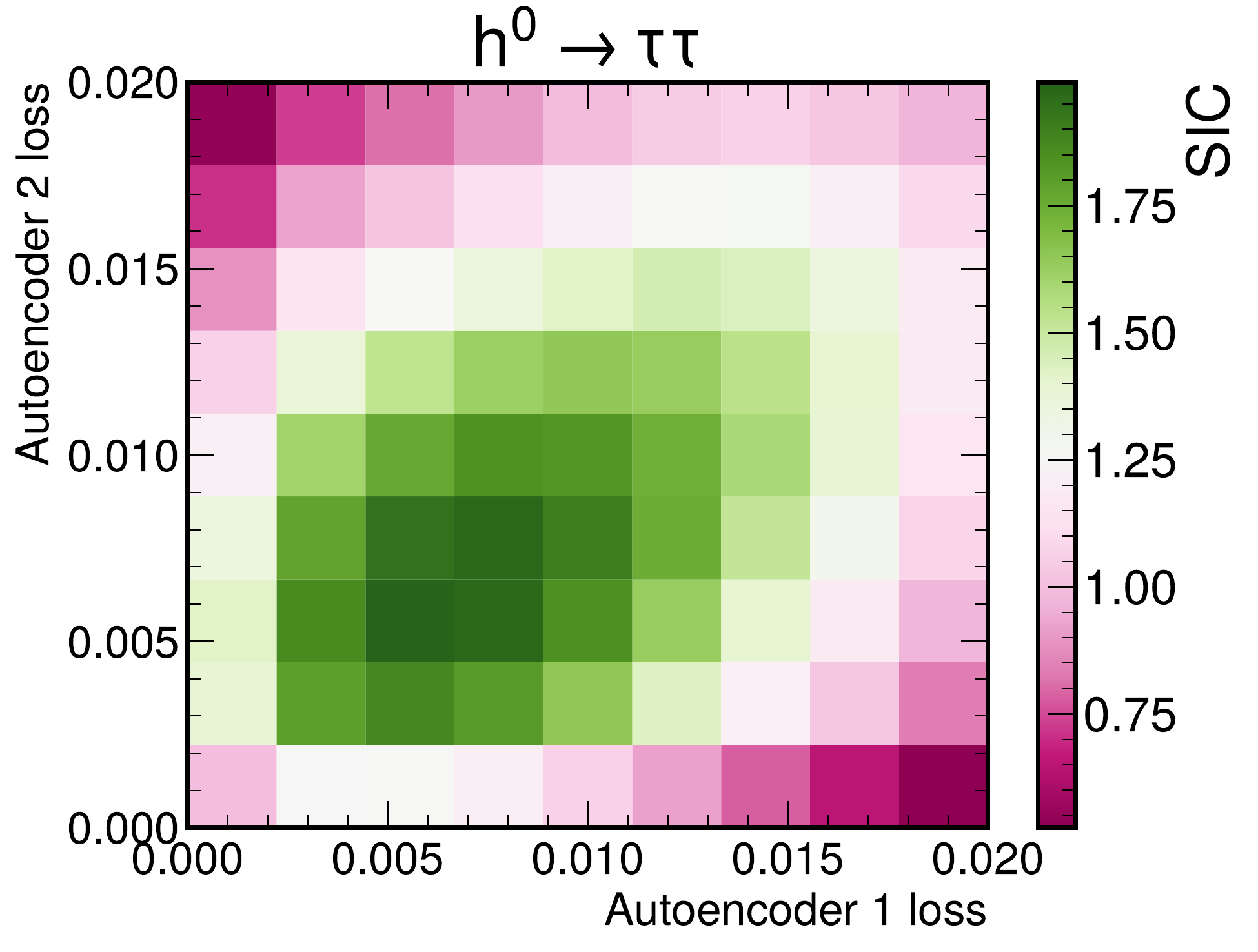}
\includegraphics[width=0.20\textwidth]{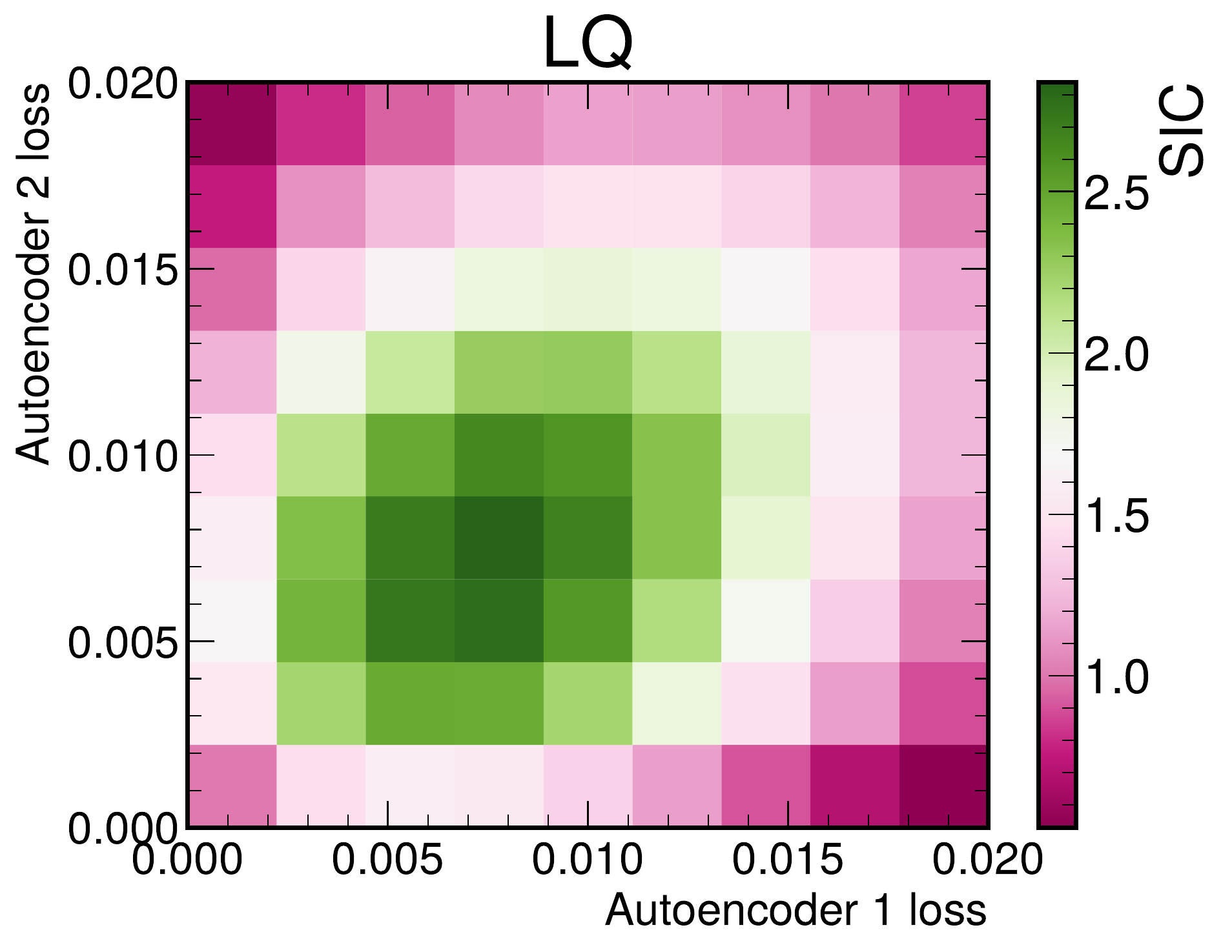}
\caption{Significance improvement characteristic for different new physics benchmark scenarios. The lower edges of each bin represents the selection threshold applied for each autoencoder loss function.}
\label{fig:sig_2d}
\end{figure}

The studies presented in this work use the ``diagonal" cut as the representative selection for the combined performance. Different choices, leading to different results, can be used when a particular new physics scenario is under study. To exemplify this difference, we show in Fig.~\ref{fig:sig_2d} the SIC curve for different selections applied to the reconstruction loss of each autoencoder. While a symmetric selection results in maximum SIC values for all benchmarks, the exact threshold resulting in maximum SIC is different for each benchmark scenario.

\end{document}